%% file: main.tex
\documentclass[10pt,twocolumn,letterpaper]{article}

\usepackage[pagenumbers]{cvpr} % To force page numbers, e.g. for an arXiv version

\usepackage[table]{xcolor}

\definecolor{lightpurple}{RGB}{243,240,252}
\definecolor{lightyellow}{RGB}{255,249,224}
\definecolor{lightblue}{RGB}{222,245,237}
\usepackage{tcolorbox}

\usepackage{cuted}
\usepackage{caption}

\usepackage{dblfloatfix}

\usepackage{capt-of}
\input{preamble}
\definecolor{cvprblue}{rgb}{0.21,0.49,0.74}
\usepackage[pagebackref,breaklinks,colorlinks,allcolors=cvprblue]{hyperref}

\def\confName{CVPR}
\def\confYear{2027}

\title{OmniMeld: Enhancing Audio-Visual Understanding via Multi-Teacher On-Policy Distillation}

\author{
\parbox{0.96\textwidth}{\centering
Daili Hua$^{1}$\thanks{These authors contributed equally.},
Yifan Dai$^{3,4*}$,
Bohan Zeng$^{1}$,
Yuran Wang$^{1}$,
Zihao Wu$^{1}$,
Chengzhuo Tong$^{1}$\\[3pt]
Guoqing Jiang$^{2}$,
Liwei Chen$^{2}$,
Zhiheng Qin$^{2}$,
Xuanping Li$^{2}$,
Wentao Zhang$^{1,4,5}$\thanks{Corresponding Author: wentao.zhang@pku.edu.cn} \\[6pt]
{\normalsize
$^{1}$Peking University \quad
$^{2}$Kuaishou Technology \quad
$^{3}$Shanghai Jiao Tong University \\[3pt]
$^{4}$ Zhongguancun Academy \quad
$^{5}$Beijing Key Laboratory of Data Intelligence and Security (Peking University)
}\\[3pt]
}
}
\begin{document}
\maketitle
\input{sec/0_abstract}    
\input{sec/1_intro}
\input{sec/2_related_works}
\input{sec/3_methods}
\input{sec/4_experiments}

\input{sec/5_conclusion}
{
    \small
    \bibliographystyle{ieeenat_fullname}
    \bibliography{main}
}

\input{sec/X_suppl}

\end{document}

%% file: sec/0_abstract.tex
\begin{abstract}

Audio-visual understanding requires perceptual grounding, temporal understanding, and semantic inference. Multi-teacher distillation can consolidate these capabilities, but direct probability matching does not explicitly capture teacher prediction changes across audio-visual conditions. We introduce OmniMeld, a multi-teacher on-policy distillation method that integrates domain expertise through modality-response-guided distribution matching. We train four teachers covering grounding and correspondence, temporal dynamics and event structure, causal and inferential reasoning, and contextual and holistic understanding. Each sample is routed to its corresponding teacher for supervision conditioned on student-generated answer prefixes. We derive transport costs from teacher prediction changes over shared candidate tokens across audio-visual conditions, using optimal transport to match student predictions to those of the teacher under complete audio-visual input. We also construct OmniMeld-20K for domain-specialized teacher training. The dataset spans four capability domains, with responses structured into evidence, reasoning, and answer sections and verified through consistency checks and media-based review. Experiments on audio-visual and general video understanding benchmarks demonstrate competitive teacher performance among similarly sized open-source models. The student improves over its base model across all evaluated benchmarks and surpasses the teachers’ base model on several benchmarks.
\end{abstract}

%% file: sec/1_intro.tex
\section{Introduction}
\label{sec:intro}
\begin{figure*}[t]
  \centering
  \includegraphics[width=\textwidth]{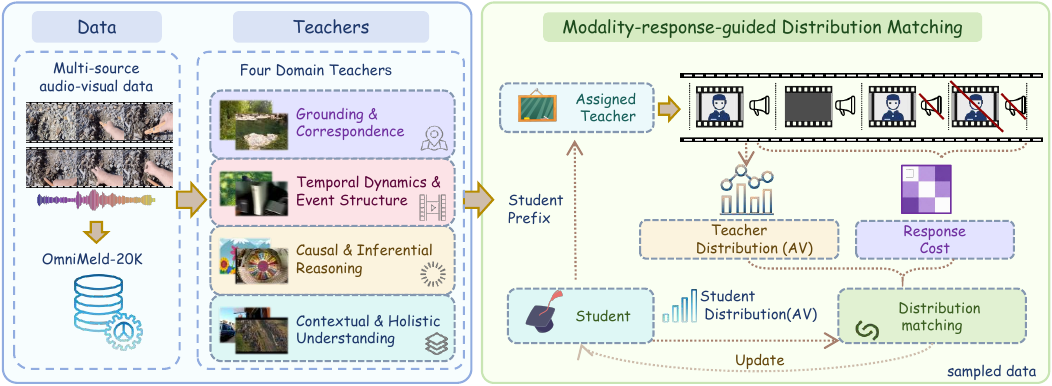}
    \caption{Overview of OmniMeld. Multi-source audio-visual data are curated into OmniMeld-20K to train four domain-specialized teachers. During modality-response-guided on-policy distillation, the assigned teacher provides domain-specific supervision on student-generated responses. Teacher prediction changes across four media conditions define transport costs for matching student and teacher distributions, with the teacher distribution under complete audio-visual input serving as the target.}
  \label{fig:teaser}
\end{figure*}

Understanding dynamic real-world scenes requires leveraging textual, auditory, and visual information. Existing omni-modal large language models provide a foundation for understanding complex multimodal scenes~\cite{team2023gemini,comanici2025gemini,pichai2025new,xu2025qwen3,team2026qwen3,team2026qwen3p8}. However, comprehensive audio-visual understanding requires complementary capabilities. For example, some questions concern objects, attributes, and audio-visual correspondence and others focus on event order and state changes. Existing research explores these capabilities with different emphases~\cite{feng2026video,li2025videochat}, yet consolidating such domain expertise into a unified model that meets diverse understanding needs remains an important challenge.

To consolidate such domain expertise, knowledge distillation offers a practical approach already adopted in foundation models~\cite{team2025gemma,meta2025llama4}. Advances in matching objectives and response utilization improve distillation~\cite{ko2024distillm,ko2025distillm}, while supervision on student-generated trajectories supports on-policy learning and cross-domain capability transfer~\cite{agarwal2024policy,ma2026mopd,xiao2026mimo}. Recent work further uses changes in visual conditions to adjust distillation weights or reconstruct target distributions~\cite{liu2026visual,zhang2026vad,li2026video,yuan2026vision}. In audio-visual settings, prediction changes induced by audio removal and interference have also been used to allocate supervision~\cite{shui2026op}. However, these methods do not jointly compare candidate tokens’ audio, visual, and interaction responses across input conditions. We therefore explore using these modality-response relationships to guide student–teacher distribution matching and support the transfer of diverse audio-visual capabilities.

Building on this idea, we introduce OmniMeld, combining domain teacher routing with modality-response-guided distribution matching, as illustrated in Fig.~\ref{fig:teaser}. To address diverse capability requirements, we train four teachers specializing in grounding and correspondence, temporal dynamics and event structure, causal and inferential reasoning, and contextual and holistic understanding. Multi-teacher on-policy distillation consolidates their expertise into a single unified student model. Each sample is assigned by domain label to a frozen teacher, with supervision conditioned on student-generated answer prefixes. We compare teacher predictions across audio-visual input conditions while holding the textual context and candidate set fixed. The resulting audio, visual, and interaction responses define transport costs between candidate tokens, guiding optimal transport matching between student and teacher distributions. This supports the transfer of domain expertise while retaining the teacher distribution under complete audio-visual input as the target.

Training these domain teachers and the unified student also requires reliable data covering diverse capability domains. We therefore construct OmniMeld-20K, separating observable evidence, reasoning, and final answers and labeling evidence by modality to make answer support traceable to specific media observations. Consistency checks, media-based verification, and constrained correction help identify unsupported or inconsistent content and correct minor factual errors, improving data reliability. Reviewed samples are organized into four capability domains for teacher training and student distillation. Experiments on audio-visual and general video understanding benchmarks show competitive performance from the domain teachers and consistent gains for the student, supporting the effectiveness of the overall pipeline in consolidating complementary expertise.

Our contributions are threefold:
\begin{itemize}
    \item We propose OmniMeld, a multi-teacher on-policy distillation framework that consolidates domain expertise through teacher routing and modality-response-guided distribution matching.
    \item We construct OmniMeld-20K to support teacher training and student distillation through structured response, consistency checks, and media-based verification.
    \item Experiments demonstrate competitive teacher performance and consistent student gains over its base model. The student also surpasses the teachers’ base model on several benchmarks.
\end{itemize}

%-------------------------------------------------------------------------

%% file: sec/2_related_works.tex
\section{Related Works}
\label{sec:related_works}

%-----------------------------------------------------------------------

\subsection{Audio-Visual Understanding}

Multimodal foundation models, including the Gemini series~\cite{team2023gemini,comanici2025gemini,pichai2025new}, Qwen-Omni series~\cite{xu2025qwen2,xu2025qwen3,team2026qwen3,team2026qwen3p8}, VideoLLaMA series~\cite{zhang2023video,zhang2024llava}, video-SALMONN 2~\cite{tang2025video}, OmniVinci~\cite{ye2026omnivinci}, Nemotron 3 Nano Omni~\cite{deshmukh2026nemotron}, and MiniCPM-o 4.5~\cite{cui2026minicpm}, have advanced audio-visual understanding. Building on these foundations, OmniVideo-R1~\cite{chen2026omnivideo} combines query-intensive grounding with modality-attentive fusion, while OmniAgent~\cite{xing2026native}, OmniReasoner~\cite{chen2026omnireasoner}, and OmniSeek~\cite{wang2026omniseek} use active perception to acquire evidence. LatentOmni~\cite{dai2026latentomni} interleaves textual reasoning with audio-visual latent states, and OmniReasoning~\cite{lin2026omnireasoning} introduces modality-factored self-distillation. While these approaches enhance perception and reasoning, we focus on consolidating complementary domain expertise into a unified student.

Alongside these modeling advances, AVDC~\cite{yan2026empowering} derives reasoning data from modality-specific and joint descriptions; OmniVideo-100K~\cite{cai2026omnivideo} generates questions from entity-anchored scripts and cross-modal clues; and Audio-Visual Flamingo~\cite{ghosh2026audio} synthesizes AV-Skills and AV-Think for perception and reasoning. Our emphasis is on structured evidence and reasoning for answer verification and correction, with verified samples organized into four capability domains for training and distillation.

\subsection{On-Policy Distillation}

On-policy distillation (OPD) obtains teacher feedback on student-generated responses. Within this framework, GKD~\cite{agarwal2024policy} supports distribution matching with different divergence objectives, while MiniLLM~\cite{gu2024minillm} optimizes reverse KL divergence through policy gradients. Multi-teacher methods further extend OPD to capability integration. MOPD~\cite{ma2026mopd} consolidates domain-specialized expertise into a shared student on student-generated trajectories and is used in MiMo-V2-Flash~\cite{xiao2026mimo} post-training.

Recent work extends OPD to multimodal understanding. CORD~\cite{hu2026cord} guides audio reasoning through text-conditioned self-distillation, and Video-OPD~\cite{li2026video} applies dense teacher feedback to temporal video grounding. For perceptual guidance, Vision-OPD~\cite{yuan2026vision} and Video-OPSD~\cite{wang2026video} use privileged image crops and evidence frames for self-distillation. Other methods exploit input variations: VA-OPD~\cite{liu2026visual} and OP-CAD~\cite{shui2026op} adjust supervision based on teacher responses to visual or audio changes, whereas VAD~\cite{zhang2026vad} reconstructs targets through visual evidence interventions. Beyond single-teacher supervision, multimodal MOPD combines complementary teachers: OPOD~\cite{zhao2026opod} coordinates text, image, and audio teachers through selective guidance, while Video-MOPD~\cite{qin2026video} integrates video capabilities through teacher routing and reliability-aware sampling.  Building on multi-teacher distillation, OmniMeld instead incorporates modality-response relationships into the distribution-matching objective. We derive transport costs between candidate tokens from changes in teacher predictions across audio-visual conditions.

%% file: sec/3_methods.tex
\section{Methods}
\label{sec:methods}

\begin{figure*}[t]
  \centering
  \includegraphics[width=\textwidth]{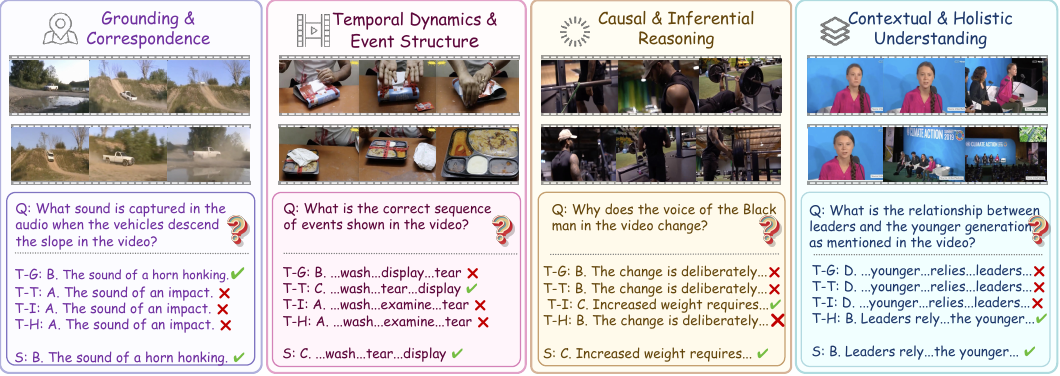}
    \caption{Examples of the four domain-specialized teachers and the distilled student. T-G, T-T, T-I, T-H, and S denote the grounding, temporal, inferential, and holistic teachers, and the student.}
  \label{fig:teachers}
\end{figure*}

%-------------------------------------------------------------------------
\subsection{Domain Specialized Teachers}

% Video question answering requires diverse capabilities, including fine-grained perception, temporal relation modeling, and high-level semantic inference. Prior work addresses complex video understanding through step-by-step processing from perception to cognition and modular decomposition into event parsing, grounding, and reasoning \cite{fei2024video, min2024morevqa}. Inspired by this capability decomposition, we organize the teacher training data into four domains according to the primary requirements of audio-visual questions: grounding and correspondence, temporal dynamics and event structure, causal and inferential reasoning, and contextual and holistic understanding. Fig.~\ref{fig:teachers} shows evaluation examples of the teachers and student.

Video question answering requires diverse capabilities, including fine-grained perception, temporal modeling, and semantic inference. Prior work decomposes complex video understanding into stages of perception, grounding, and reasoning~\cite{fei2024video, min2024morevqa}. Motivated by this modular perspective, we organize teacher training data according to distinct audio-visual capability requirements. We define four domains: grounding and correspondence, temporal dynamics and event structure, causal and inferential reasoning, and contextual and holistic understanding. Fig.~\ref{fig:teachers} shows evaluation examples of the teachers and student.

\noindent \textbf{Grounding and Correspondence Teacher:}
% This expert focuses on establishing correspondences between questions and observable audio-visual content. Its scope includes fine-grained entity recognition, spatial localization, matching sounds to their visual sources, and audio-visual event matching. The central objective is to identify the objects, attributes, or local facts referenced by a question and associate them with specific perceptual evidence.
This expert establishes correspondences between questions and observable audio-visual content. Its scope includes fine-grained entity recognition, spatial localization, matching sounds to their visual sources, and audio-visual event matching. It identifies the relevant objects, attributes, or local facts referenced by a question and associates them with specific perceptual evidence.

\noindent \textbf{Temporal Dynamics and Event Structure Teacher:}
% This expert focuses on how events are organized and unfold over time. Its scope includes event ordering, before-and-after relations, state changes, process tracking, and temporal localization. It emphasizes temporal relationships across different time points or events. 
This expert focuses on how events are organized and unfold over time. Its scope includes event ordering, before-and-after relations, state changes, process tracking, and temporal localization. It emphasizes temporal relationships across different time points or events to understand how actions and states evolve throughout a sequence.

\noindent \textbf{Causal and Inferential Reasoning Teacher:}
% This expert uses audio-visual evidence to infer conclusions that are not directly presented. Its scope includes causal explanation, intention inference, future-event prediction, and hypothetical or counterfactual analysis. The central objective is to establish justified connections between observable facts and inferred conclusions, while ensuring that the certainty expressed is appropriate to the available evidence.
This expert uses audio-visual evidence to draw inferences beyond directly observable facts. Its scope includes causal explanation, intention inference, future-event prediction, and hypothetical or counterfactual analysis. It establishes justified connections between observable facts and inferred conclusions, ensuring that inferences remain supported by the available evidence.

\noindent \textbf{Contextual and Holistic Understanding Teacher:}
% This expert integrates cues distributed throughout an audio-visual sequence to form a coherent understanding of its overall content and context. Its scope includes theme identification, content summarization, contextual interpretation, and semantic comparison across segments. The central objective is to combine local information into a global interpretation that supports questions requiring an understanding of the broader context.
This expert integrates audio-visual cues throughout a video to form a coherent understanding of its overall content and context. Its scope includes theme identification, content summarization, contextual interpretation, and semantic comparison across different parts. It addresses questions that require interpreting specific details within this broader context.

% During student distillation, each example is routed to a single teacher according to its assigned domain label. The selected teacher provides supervision on the student-generated answer prefixes, while all teacher parameters remain frozen. Knowledge from the four domains is consolidated through updates to the shared student model.
% During distillation, each example is routed to a frozen teacher according to its domain label. The selected teacher provides supervision conditioned on student-generated answer prefixes, allowing knowledge from the four domains to be consolidated into a shared student model.
During distillation, each example is routed to a frozen teacher according to its domain label. The selected teacher provides domain-specific guidance on student-generated responses, allowing complementary knowledge from the four domains to be consolidated into a shared student model.
% --------------------------------------------------------------------------------
% \subsection{Audio-Visual Response Guided Transport for On-Policy Distillation}

% \textbf{Base Distillation Framework:}
% \textbf{}

\subsection{Modality-Response-Guided On-Policy Distillation}
\label{sec:art_opd}

\textbf{Preliminaries: Multi-Teacher On-Policy Distillation.}
We adopt a GKD-based multi-teacher on-policy distillation framework. Given an audio-visual input \(x=(q,a,v)\), the student generates an answer \(y=(y_1,\ldots,y_L)\), and a frozen teacher \(T=T_{k(x)}\) is selected according to the task domain. At each position \(t\), the teacher supervises the student’s next-token distribution conditioned on the same input and student-generated prefix \(y_{<t}\).

We use the teacher's top-$K$ tokens as the shared candidate set $\mathcal C_t$ to reduce computation. Let $z^S_{t,j}$ and $\ell^{T,AV}_{t,j}$ denote the student logit and teacher log-probability for the $j$-th candidate. We normalize both distributions over this set using temperature $\tau$:

\[
p^S_{t,j}
=
\frac{\exp(z^S_{t,j}/\tau)}
{\sum_{h=1}^{K}\exp(z^S_{t,h}/\tau)},
\qquad
p^T_{t,j}
=
\frac{\exp(\ell^{T,AV}_{t,j}/\tau)}
{\sum_{h=1}^{K}\exp(\ell^{T,AV}_{t,h}/\tau)}.
\]

We choose reverse KL as the base GKD objective:

\[
\mathcal L_{\mathrm{GKD}}^t
=
D_{\mathrm{KL}}(p^S_t\|p^T_t)
=
\sum_{j=1}^{K}p^S_{t,j}
\log\frac{p^S_{t,j}}{p^T_{t,j}}.
\]

Gradients are computed only with respect to the student distribution, while teacher scores are held fixed. Knowledge from the assigned teachers is consolidated through updates to the shared student model.

%------------------------------------3.3.2

\noindent \textbf{Modality Response Modeling.}
The base objective compares student and teacher probabilities but does not explicitly account for differences in candidate responses to audio-visual information. We therefore construct a transport cost from the assigned teacher's predictions under different input conditions.

For a preselected subset of training examples, we construct four views: complete audio-video $AV$; an audio-preserving view $A$ with black video frames; a visual view $V$ with audio disabled; and a media-free view $\varnothing$. All views retain the same question and student-generated answer prefix. The candidate set is defined by the assigned teacher's top-$K$ predictions under $AV$ and remains fixed when the same teacher scores candidates across all four views.

Omitting the position index $t$, the teacher score for candidate $w_j$ under view $c$ is
\begin{equation}
\begin{aligned}
\ell_j^c&=\log p_{T_{k(x)}}(w_j\mid x^c,y_{<t}),\\
c&\in\{AV,A,V,\varnothing\},
\end{aligned}
\label{eq:art_teacher_views}
\end{equation}
where $x^c$ denotes the corresponding input view. Using teacher log-probabilities at temperature $1$, we define a modality-response vector:
\begin{equation}
r_j=\begin{bmatrix}
\ell_j^A-\ell_j^\varnothing\\
\ell_j^V-\ell_j^\varnothing\\
\ell_j^{AV}-\ell_j^A-\ell_j^V+\ell_j^\varnothing
\end{bmatrix}.
\label{eq:art_response}
\end{equation}
The components represent audio, visual, and audio-visual interaction responses with textual context held constant. The response discrepancy $D_{jh}$, normalization scale $s$, and transport cost $C_{jh}$ are
\begin{equation}
\begin{aligned}
D_{jh}&=\|r_j-r_h\|_2^2,\\
s&=\max\left(1,\frac{1}{K(K-1)}\sum_{u\ne v}D_{uv}\right),\\
C_{jh}&=\mathbf{1}[j\ne h]+\frac{D_{jh}}{s}.
\end{aligned}
\label{eq:art_cost}
\end{equation}
Here, $j,h,u,v$ index the shared candidates. Larger response discrepancies incur higher transport costs. The indicator $\mathbf{1}[j\ne h]$ prevents free transport between distinct tokens, while the lower bound on $s$ avoids amplifying negligible response differences.

%----------------------------------------
\begin{figure*}[t]
  \centering
  \includegraphics[width=\textwidth]{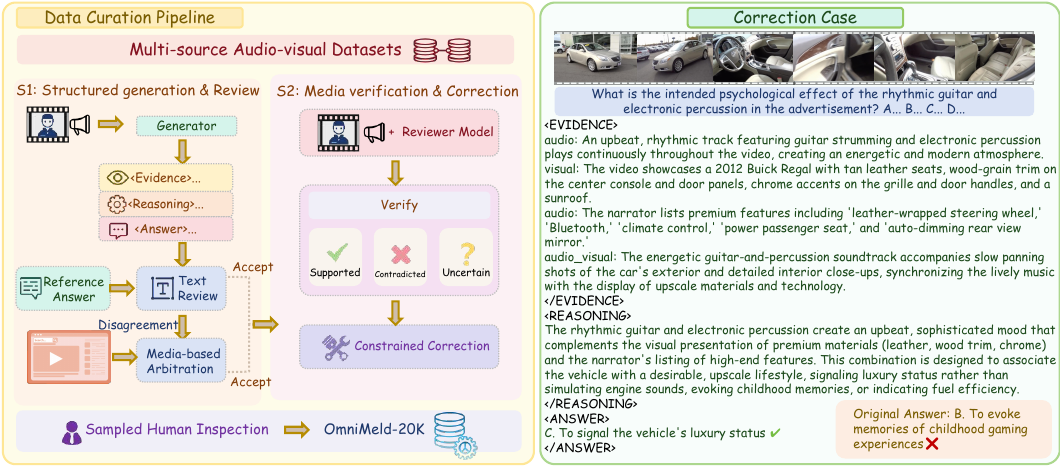}
    \caption{Data curation pipeline for OmniMeld-20K, combining structured generation, consistency review, media-based verification, constrained correction, and sampled human inspection. An illustrative example on the right side shows answer correction supported by the data curation pipeline.}
  \label{fig:data pipeline}
\end{figure*}

\paragraph{Optimal Transport-Based Distribution Matching.}
We retain the teacher's complete-input distribution $p^T$ as the target and compare it with $p^S$ using the response-based cost. For distributions $\mu,\nu$, the feasible transport plans are
\begin{equation}
\Pi(\mu,\nu)=\left\{\Gamma\ge0:
\Gamma\mathbf{1}=\mu,\;
\Gamma^\top\mathbf{1}=\nu\right\},
\label{eq:art_transport_plans}
\end{equation}
where $\Gamma$ is a $K\times K$ transport matrix and $\mathbf{1}$ is the all-ones vector. We define entropy-regularized optimal transport as
\begin{equation}
\begin{aligned}
F_\varepsilon(\mu,\nu;C)
=\min_{\Gamma\in\Pi(\mu,\nu)}\Bigg\{&
\sum_{j,h}\Gamma_{jh}C_{jh}\\
&+\varepsilon\sum_{j,h}\Gamma_{jh}(\log\Gamma_{jh}-1)
\Bigg\},
\end{aligned}
\label{eq:art_regularized_ot}
\end{equation}
where $\varepsilon$ controls entropy regularization. We solve these problems using Sinkhorn~\cite{cuturi2013sinkhorn} iterations and adopt the debiased loss~\cite{feydy2019interpolating}
\begin{equation}
\begin{aligned}
\mathcal L_{\mathrm{OT}}^t
={}&F_\varepsilon(p^S_t,p^T_t;C_t)\\
&-\frac12F_\varepsilon(p^S_t,p^S_t;C_t)\\
&-\frac12F_\varepsilon(p^T_t,p^T_t;C_t).
\end{aligned}
\label{eq:art_sinkhorn_divergence}
\end{equation}
All three terms use the same teacher-derived cost matrix $C_t$ at position $t$. The self-transport terms remove entropic self-bias, yielding zero loss when the student and teacher distributions coincide.

Let $m_i\in\{0,1\}$ denote a predefined sample-selection indicator. Selected samples are trained using optimal transport (OT) at all valid answer positions, while the remaining samples retain GKD:
\begin{equation}
\begin{aligned}
\mathcal L=\frac{1}{|\mathcal B|}
\sum_{i\in\mathcal B}\frac{1}{L_i}
\sum_{t=1}^{L_i}\Big[
&(1-m_i)\mathcal L_{\mathrm{GKD}}^{i,t}\\
&+m_i\mathcal L_{\mathrm{OT}}^{i,t}\Big].
\end{aligned}
\label{eq:art_final_objective}
\end{equation}
% Here, $\mathcal B$ is the training batch and $L_i$ counts valid answer tokens for sample $i$. This averaging matches our configuration with one sample per microbatch. Approximately $20\%$ of samples are selected in advance for OT; this proportion is not an additional loss coefficient. Given comparable loss scales in preliminary diagnostics, we apply no additional scaling. We use $K=64$, $\tau=0.9$, and $\varepsilon=0.1$.

Here, $\mathcal B$ is the training batch and $L_i$ counts valid answer tokens for sample $i$. We apply no additional loss scaling.

Teacher scores and transport costs are treated as constants, and gradients update only the student under complete audio-visual input. Additional views are used solely for teacher scoring during training; at inference, the student directly generates answers from the complete input.

\begin{figure}[t]
    \centering
    \includegraphics[width=\linewidth]{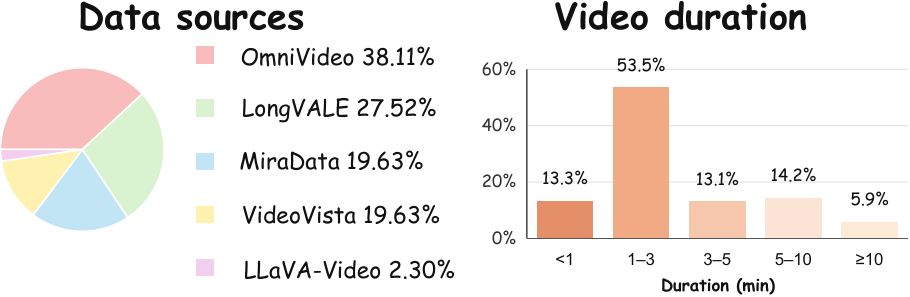}
    \caption{OmniMeld-20K composition across five source datasets on the left and video duration distribution on the right.}
    \label{fig:dataset compositon}
\end{figure}

\begin{table*}[t]
    \caption{Performance comparison across four audio-visual understanding benchmarks. Best results among closed-source reference models and within each open-source model size group are highlighted in bold.}
  \label{tab:benchmark_comparison}
  \centering
  \small
  \setlength{\tabcolsep}{3pt}
  \newcommand{\AudioDurationTable}{%
  \begin{tabular}{lccccc}
    \toprule
    Model & Size & DailyOmni & OmniVideoBench & WorldSense & LVOmniBench \\
    \midrule
    \rowcolor{lightyellow}
    \multicolumn{6}{c}{\textit{Closed-source Models}} \\
    Gemini-2.5-Pro      & --   &81.4 & 58.9 & 64.6 & -- \\ 
    Gemini-3.0-Pro       & --  & 81.1 & \textbf{61.8} & \textbf{66.4} & \textbf{65.8} \\
    Qwen3.5-Omni-Flash  & --  & \textbf{81.8} & --   & 57.9 & --   \\
    \midrule
    \rowcolor{lightyellow}
    \multicolumn{6}{c}{\textit{Open-source Models}} \\
    Uni-MoE-2.0-Omni    & 30B    & 64.3 & 38.6 & --   & --   \\
    Qwen3-Omni-Instruct & 30B    & 71.9 & 43.6 & 55.1 & 35.8 \\
    Qwen3-Omni-Thinking & 30B    & 73.6 & 39.6 & 52.7 & 31.9 \\
    Nemotron-3-Nano-Omni    & 30B    & 74.5 & --   & 55.2 & --   \\
    OmniVideo-R1        & 30B    & \textbf{82.8} & 44.8 & \textbf{65.8} & -- \\
    OmniSeek          & 30B    & 80.0 & \textbf{47.7} & 62.4 & \textbf{44.2} \\
    \midrule
    VideoLLaMA2          & 7B     & 35.2 & 29.2 & 25.4 & 27.2 \\
    VITA-1.5            & 7B     & 44.7 & 30.5 & 36.9 & --   \\
    HumanOmniV2         & 7B     &58.5 & 30.5 & 47.1 & 32.3\\
    Baichuan-Omni-1.5   & 7B     &50.0 & 30.7 & 43.3 & 32.8\\
    Qwen2.5-Omni        & 7B     & 61.3 & 30.4 & 42.4 & 33.4 \\
    OmniReasoner         & 7B     & 64.2 & 34.8 & 46.7 & 35.4 \\
    OmniAgent           & 7B     & 64.8 & 37.1 & 47.2 & --   \\
    OmniVinci           & 7B     & 66.5 & --   & 48.2 & --   \\
    LatentOmni          & 7B     & 67.4 & 35.4 & \textbf{48.9}\textsuperscript{*} & 35.1 \\
    \rowcolor{lightpurple}
    T-Grounding        & 7B     & 68.9 & 35.8 & 44.1 & 34.5 \\
    \rowcolor{lightpurple}
    T-Temporal         & 7B     & \textbf{71.0} & 39.0 & 45.8 & 35.8 \\
    \rowcolor{lightpurple}
    T-Inferential           & 7B     & 67.1 & \textbf{39.1} & 46.5 & 36.5 \\
    \rowcolor{lightpurple}
    T-Holistic        & 7B     & 68.5 & \textbf{39.1} & 46.5 & \textbf{36.7} \\
    \midrule
    Qwen2.5-Omni        & 3B     & 60.1 & 31.8 & 41.3 & 32.3 \\
    \rowcolor{lightpurple}
    OmniMeld(ours)                & 3B     & \textbf{66.2} & \textbf{34.7} & \textbf{43.7} & \textbf{33.1} \\
    \bottomrule
  \end{tabular}
  }
  \setlength{\tabcolsep}{0pt}
  \sbox{0}{\AudioDurationTable}
  % Nine columns have eight internal gaps, each containing two tabcolsep spaces.
  \setlength{\tabcolsep}{\dimexpr(\textwidth-\wd0)/12\relax}
  \AudioDurationTable
  \par\vspace{2pt}
    \begin{minipage}{\linewidth}
    \footnotesize
    \textsuperscript{*}Result reported in~\cite{dai2026latentomni}. The corresponding base model scores 45.4 on WorldSense, compared with 42.4 in our re-evaluation. 
    \end{minipage}
\end{table*}

\subsection{Multi Source Audio-Visual Data Curation}
\label{subsec: data curation}

% Existing work on audio-visual understanding uses cross-modal evidence chains to construct QA data ~\cite{cai2026omnivideo} or aligns latent states with perceptual features to support implicit reasoning ~\cite{dai2026latentomni}. We focus on the verifiability of training data by explicitly separating observable facts, evidence-based reasoning, and final answers into three distinct sections. Each observation is labeled with its modality source, reasoning draws on the listed evidence, and the answer is supported by both. This organization enables separate checks of evidence support, reasoning consistency, and answer correctness, while facilitating error localization and constrained correction. Based on this design, we refine and verify audio-visual data from multiple sources to provide reliable supervision for subsequent training. Fig.~\ref{fig:data pipeline} illustrates the overall pipeline and a correction case.

% Existing work constructs audio-visual training data using cross-modal evidence chains~\cite{cai2026omnivideo} or supports implicit reasoning through latent–perceptual alignment~\cite{dai2026latentomni}. We focus on data verifiability by separating observable evidence, reasoning, and final answers. Evidence items are labeled by modality, and reasoning and answers are grounded in the listed evidence. This structure facilitates verification and correction of multi-source audio-visual data for training. Fig.~\ref{fig:data pipeline} illustrates the pipeline and a correction case.

Existing work constructs audio-visual training data through cross-modal evidence chains~\cite{cai2026omnivideo} or supports implicit reasoning through latent–perceptual alignment~\cite{dai2026latentomni}. We emphasize data verifiability by separating modality-labeled evidence, reasoning, and final answers, with reasoning and answers grounded in the evidence. This structure supports verification and correction of multi-source training data.

% Using this approach, we combine and process existing audio-visual resources to construct OmniMeld-20K, comprising approximately 20K training examples. Fig. ~\ref{fig:dataset compositon} summarizes the dataset composition and video duration distribution. Videos and candidate questions originate from
% OmniVideo-100K~\cite{cai2026omnivideo}, MiraData~\cite{ju2024miradata}, LongVALE~\cite{geng2025longvale}, VideoVista~\cite{li2024videovista}, and LLaVA-Video~\cite{zhang2024llava}. The questions associated with MiraData and LongVALE videos are sourced from AV-Skills, the dataset introduced in Audio-Visual Flamingo~\cite{ghosh2026audio}, while the selected VideoVista and LLaVA-Video questions come from the resources curated in OmniVideo-R1~\cite{chen2026omnivideo}. Finally, we manually inspect a subset of the data to assess its quality. The curation process comprises two stages, detailed below.

Using this approach, we combine and process existing audio-visual resources to construct OmniMeld-20K, comprising approximately 20K training examples. Fig.~\ref{fig:dataset compositon} summarizes its composition and video duration distribution. Videos and candidate questions originate from OmniVideo-100K~\cite{cai2026omnivideo}, MiraData~\cite{ju2024miradata}, LongVALE~\cite{geng2025longvale}, VideoVista~\cite{li2024videovista}, and LLaVA-Video~\cite{zhang2024llava}. Questions for MiraData and LongVALE videos come from AV-Skills in Audio-Visual Flamingo~\cite{ghosh2026audio}, while VideoVista and LLaVA-Video questions come from resources curated in OmniVideo-R1~\cite{chen2026omnivideo}. Fig.~\ref{fig:data pipeline} illustrates the two-stage curation process and a correction case. We detail both stages below.

\noindent \textbf{Stage 1: Structured Generation and Consistency Review.}
An omnimodal model (Qwen3.8-Omni-Flash~\cite{team2026qwen3p8}) receives the complete audio-visual input and question, generating \texttt{<EVIDENCE>}, \texttt{<REASONING>}, and \texttt{<ANSWER>} sections. Evidence records observable facts labeled as \texttt{visual}, \texttt{audio}, or \texttt{audio\_visual}, the last indicating observations requiring both modalities, such as synchronization or correspondence. Reasoning is expected to follow the listed evidence without adding observations, and the answer section provides the final response. An independent text model (GPT-5.6 Luna~\cite{openai2026gpt56luna}) then checks evidence–reasoning consistency and semantic agreement with reference answers. A third-party model with media access (Gemini 3.6 Flash~\cite{google2026gemini36flash}) adjudicates substantial disagreements identified during this review. Failed samples are regenerated and discarded after repeated failures.

\noindent \textbf{Stage 2: Media-Based Evidence Verification and Constrained Correction.}
% After initial response generation, the same model assumes a reviewer role and systematically reassesses each response against the original media. It classifies individual evidence items as supported, contradicted, or uncertain, while evaluating reasoning soundness and answer correctness. Corrections are restricted to local errors in counts, timestamps, quotations, and entities, preserving the original scope of observation and modality attribution without introducing new claims. This layered verification and constrained correction process improves consistency between annotations and media content, providing reviewed supervision for subsequent training.
%------
% After initial response generation, the same model assumes a reviewer role and reassesses each response against the original media. It classifies evidence items as supported, contradicted, or uncertain, while evaluating reasoning soundness and answer correctness. Corrections are limited to minor factual errors, preserving the original scope of observation and modality attribution without introducing new claims. This verification and constrained correction process improves consistency between responses and media content, providing reviewed supervision for training.
After generation, the same model assumes a reviewer role and reassesses responses against the media. It classifies evidence as supported, contradicted, or uncertain and evaluates reasoning soundness and answer correctness. Corrections only address minor factual errors, preserving observation scope and modality attribution without introducing new claims. This verification and constrained correction improve consistency between responses and media content, providing training supervision. Finally, we manually inspect a subset to assess quality.

%% file: sec/4_experiments.tex
\section{Experiments}
\label{sec:experiments}

\begin{table*}[t]
  \centering
  \small
  \setlength{\tabcolsep}{3pt}
    \caption{Performance on OmniVideoBench across audio categories and video duration groups. Comparisons focus on teachers against similarly sized models and the student against its base model. Best results within each group are highlighted in bold.}
  \label{tab:omnivideobench}
  % Measure the table without padding, then distribute the remaining width.
  \newcommand{\AudioDurationTable}{%
  \begin{tabular}{lcccccccc}
    \toprule
    Method & \multicolumn{3}{c}{Audio Type} & \multicolumn{4}{c}{Video Duration} & Avg. \\
    \cmidrule(lr){2-4}\cmidrule(lr){5-8}
    & Music & Sound & Speech & (0,1] min & (1,5] min & (5,10] min & (10,30] min & \\
    \midrule
    \rowcolor{lightyellow}
    \multicolumn{9}{c}{\textit{Closed-source Models}} \\
    Gemini-2.5-Pro & 38.5 & \textbf{57.7} & \textbf{61.7} & 57.8 & \textbf{64.4} & \textbf{55.0} & \textbf{55.9} & \textbf{58.9} \\
    \midrule
    \rowcolor{lightyellow}
    \multicolumn{9}{c}{\textit{Open-source Models}} \\
    VideoLLaMA2 & 26.4 & 30.7 & 29.3 & 32.0 & 28.2 & 29.6 & 28.3 & 29.2 \\
    VITA-1.5 & 25.3 & 28.6 & 31.5 & 31.3 & 27.4 & 30.6 & 34.0 & 30.5 \\
    HumanOmniV2 & 20.9 & 31.1 & 31.6 & 36.6 & 29.4 & 29.6 & 29.3 & 30.5 \\
    Baichuan-Omni-1.5 & 24.2 & 31.3 & 31.4 & 28.9 & 31.8 & 28.4 & 32.4 & 30.7 \\
    Qwen2.5-Omni-7B & 25.3 & 23.8 & 32.3 & 42.7 & 30.0 & 27.8 & 25.3 & 30.4 \\
    OmniReasoner & 34.1 & 29.3 & 36.0 & - & - & 31.9 & 36.6 & 34.8\\
    LatentOmni & 33.3 & 30.2 & 36.7 & 45.2 & 33.2 & 33.3 & 34.0 & 35.4\\
    \rowcolor{lightpurple}
    T-Grounding & 27.8 & 30.2 &  37.9& 42.9 & 36.0 & 35.0 & 32.0 & 35.8 \\
    \rowcolor{lightpurple}
    T-Temporal & \textbf{34.4} & 31.8 & \textbf{40.9} & \textbf{48.0} & \textbf{39.5} & 37.6 & 33.6 & 39.0 \\
    \rowcolor{lightpurple}
    T-Inferential   & 31.9 & \textbf{35.6} & 40.7 & 46.8 & 39.1 & 38.5 & 35.0 & \textbf{39.1} \\
    \rowcolor{lightpurple}
    T-Holistic & 32.6 & 34.0 & \textbf{40.9} & 45.2 & 37.3 & \textbf{38.9} & \textbf{37.7} & \textbf{39.1} \\
    \midrule
    Qwen2.5-Omni-3B & 26.7 & 32.2 & 32.4 & 40.0 & 31.4 & 29.5 & 29.2 & 31.8 \\
    \rowcolor{lightpurple}
    OmniMeld(ours)            & \textbf{28.6} & 30.8 & \textbf{36.2} & \textbf{40.9} & \textbf{34.5}& \textbf{33.9} & \textbf{31.9} & \textbf{34.7} \\
    \bottomrule
  \end{tabular}%
  }
  \setlength{\tabcolsep}{0pt}
  \sbox{0}{\AudioDurationTable}
  % Nine columns have eight internal gaps, each containing two tabcolsep spaces.
  \setlength{\tabcolsep}{\dimexpr(\textwidth-\wd0)/18\relax}
  \AudioDurationTable
\end{table*}

% Video-MME: single-column version.
\begin{table}[t]
    \caption{Performance comparison on the general video understanding benchmark Video-MME using audio-visual inputs without subtitles. Best results are highlighted in bold.}
  \label{tab:videomme}
  \centering
  \small
  \newcommand{\VideoMMETable}{%
  \begin{tabular}{lcccc}
    \toprule
    Method & Short & Medium & Long & Overall \\
    \midrule
    \cellcolor{lightblue}Qwen2.5-Omni-7B & 76.7 & 66.8 & 54.7 & 66.0 \\
    OmniReasoner & - & - & - & 65.4 \\
    OmniAgent & - & - & \textbf{59.6} & 67.8 \\
    \rowcolor{lightpurple}
    T-Grounding & 76.9 & 66.7 & 55.9 & 66.5 \\
    \rowcolor{lightpurple}
    T-Temporal & \textbf{79.2} & 69.0 & 56.2 & 68.1 \\
    \rowcolor{lightpurple}
    T-Inferential & 77.7 & \textbf{69.1} & 57.8 & \textbf{68.2} \\
    \rowcolor{lightpurple}
    T-Holistic & 78.4 & 68.8 & 57.4 & \textbf{68.2} \\
    \midrule
    \cellcolor{lightblue}Qwen2.5-Omni-3B & 74.6 & 62.1 & 49.9 & 62.2 \\
    \rowcolor{lightpurple}
    OmniMeld(ours) & \textbf{75.3} & \textbf{65.1} & \textbf{51.3} & \textbf{63.9} \\
    \bottomrule
  \end{tabular}%
  }
  \setlength{\tabcolsep}{0pt}
  \sbox{0}{\VideoMMETable}
  % Five columns, including padding at both outer edges: 2 * 5 = 10.
  \setlength{\tabcolsep}{\dimexpr(\columnwidth-\wd0)/10\relax}
  \VideoMMETable
\end{table}

\begin{table*}[ht]
  \caption{ Ablation study on training strategy. Best results are highlighted in bold.}
  \label{tab:ablation}
  \centering
  \small
  \newcommand{\AudioDurationTable}{%
  \begin{tabular}{lccccc}
    \toprule
    Method & Daily-Omni& OmniVideoBench & WorldSense & LVOmniBench & Avg.\\
    \midrule
    \cellcolor{lightblue} Qwen2.5-Omni-3B & 60.1 & 31.8 & 41.3 & 32.3 & 41.4\\
    + SFT Warmup  & 63.1 & 33.6 & 43.6 & \textbf{33.1} & 43.4 \\
    + SFT Warmup + MOPD  & 65.7 & 33.6 & 43.5 & \textbf{33.1} & 44.0 \\
    \rowcolor{lightpurple}
    + SFT Warmup + MOPD w/ OT & \textbf{66.2} & \textbf{34.7} & \textbf{43.7} & \textbf{33.1} & \textbf{44.4} \\
    \bottomrule
  \end{tabular}
  }
  \setlength{\tabcolsep}{0pt}
  \sbox{0}{\AudioDurationTable}
  % Nine columns have eight internal gaps, each containing two tabcolsep spaces.
  \setlength{\tabcolsep}{\dimexpr(\textwidth-\wd0)/12\relax}
  \AudioDurationTable
\end{table*}

\noindent \textbf{Baselines:}
We primarily compare our teachers with open-source models of a comparable 7B parameter scale, including VideoLLaMA2~\cite{cheng2024videollama}, VITA-1.5~\cite{fu2026vita}, HumanOmniV2~\cite{yang2025humanomniv2}, Baichuan-Omni-1.5~\cite{li2025baichuan}, Qwen2.5-Omni~\cite{xu2025qwen2}, OmniReasoner~\cite{chen2026omnireasoner}, OmniAgent~\cite{xing2026native}, OmniVinci~\cite{ye2026omnivinci} and  LatentOmni~\cite{dai2026latentomni}. For the 3B student, we use the Qwen2.5-Omni-3B base model as a direct baseline. We additionally report results from larger open-source models, predominantly at the 30B scale, including Uni-MoE-2.0~\cite{li2025uni}, Qwen3-Omni-Instruct~\cite{xu2025qwen3}, Qwen3-Omni-Thinking~\cite{xu2025qwen3}, Nemotron-3-Omni~\cite{deshmukh2026nemotron}, OmniVideo-R1~\cite{chen2026omnivideo}, OmniSeek~\cite{wang2026omniseek}, alongside closed-source models such as Gemini-2.5-Pro~\cite{comanici2025gemini}, Gemini-3-Pro~\cite{pichai2025new}, and Qwen3.5-Omni-Flash~\cite{team2026qwen3}, to provide broader performance context. Results for comparison models are taken from their respective publications, while the Qwen2.5-Omni baselines, our teachers, and our student are evaluated locally.

\noindent \textbf{Benchmarks:}
We evaluate our models on joint audio-visual understanding and general video understanding. For joint audio-visual understanding, we use Daily-Omni~\cite{zhou2025daily}, OmniVideoBench~\cite{li2026omnivideobench}, WorldSense~\cite{hong2026worldsense}, and LVOmniBench~\cite{tao2026lvomnibench}, which evaluate audio-visual reasoning in everyday scenes, real-world omnimodal understanding, synergistic audio-visual reasoning, and long-form audio-visual understanding, respectively. Beyond overall accuracy, we analyze performance across the audio-type and video-duration subsets of OmniVideoBench. For general video understanding, we use Video-MME~\cite{fu2025video} with audio input and without subtitles, reporting overall accuracy alongside results on the short, medium, and long video subsets.

\subsection{Experiment Settings}
\noindent \textbf{Implementation Details:}
% We initialize four domain-specialized teachers from Qwen2.5-Omni-7B~\cite{xu2025qwen2} and the student from Qwen2.5-Omni-3B~\cite{xu2025qwen2}. Each teacher undergoes supervised fine-tuning (SFT) on approximately 5K samples from its corresponding domain in OmniMeld-20K, curated in Sec.~\ref{subsec: data curation}. The student is first warmed up through SFT on 2K examples, followed by on-policy distillation on 3,840 samples. Both sets are randomly drawn from OmniMeld-20K used to train the teachers. In our implementation, we use reverse KL divergence over the teacher's Top-64 candidates as the base distillation objective and replace it with the OT loss for a fixed subset comprising 20\% of the training samples.
We initialize four domain-specialized teachers from Qwen2.5-Omni-7B~\cite{xu2025qwen2} and the student from Qwen2.5-Omni-3B~\cite{xu2025qwen2}. Each teacher undergoes supervised fine-tuning (SFT) on approximately 5K samples from its corresponding domain in OmniMeld-20K, curated in Sec.~\ref{subsec: data curation}. For student training, we randomly draw two subsets from the same dataset, containing 2K and 3,840 samples, respectively. The former is used for SFT warmup, and the latter for subsequent on-policy distillation under the domain-specialized teachers. In our implementation, we use reverse KL divergence over the teacher's top-$K$ candidates as the base distillation objective and replace it with the OT loss for a preselected subset comprising 20\% of the distillation samples. Specifically, we set the candidate set size to $K=64$, the shared distribution temperature to $\tau=0.9$, and the entropy regularization strength to $\varepsilon=0.1$.

\subsection{Main Results}

\textbf{Comparison between Teachers and Open-source Models.} Tab.~\ref{tab:benchmark_comparison} shows that our domain teachers achieve competitive performance against similarly sized open-source models while consistently improving over the locally evaluated Qwen2.5-Omni-7B baseline across all four benchmarks. The Temporal Dynamics and Event Structure Teacher (T-Temporal) achieves 71.0 on DailyOmni, improving over the base model by 9.7 percentage points. The Causal and Inferential Reasoning Teacher (T-Inferential) and the Contextual and Holistic Understanding Teacher (T-Holistic) reach 39.1 on OmniVideoBench, a gain of 8.7 percentage points. T-Holistic achieves 36.7 on LVOmniBench, improving by 3.3 percentage points. On WorldSense, T-Inferential and T-Holistic score 46.5, exceeding the baseline by 4.1 percentage points. These gains support the effectiveness of domain-specialized teacher training.

% \noindent \textbf{Comparison between Different Teachers.} The teachers show strengths reflecting their training focus. T-Temporal emphasizes event ordering and temporal relationships, performing well on DailyOmni, while T-Inferential and T-Holistic show more balanced performance through evidence-based inference and contextual integration. The Grounding and Correspondence Teacher (T-Grounding) focuses on objects, attributes, and audio-visual correspondence, with strengths that may be less apparent in aggregate scores. As shown in Fig.~\ref{fig:teachers}, T-Grounding and the student produce the correct answer, whereas the other teachers answer incorrectly. This example illustrates the teachers’ complementary expertise and the value of perceptual supervision.
\noindent \textbf{Comparison between Different Teachers.} The teachers show distinct strengths reflecting their training focus. T-Temporal emphasizes event ordering and temporal relationships, performing well on DailyOmni, while T-Inferential and T-Holistic show more balanced performance, consistent with their focus on inference and contextual understanding. The Grounding and Correspondence Teacher (T-Grounding) focuses on objects, attributes, and audio-visual correspondence, with strengths that may be less apparent in aggregate scores. Fig.~\ref{fig:teachers} compares teacher and student responses across all four capability domains. These examples illustrate the teachers’ complementary expertise and the student’s ability to address questions spanning these domains.

\noindent \textbf{Student Analysis.} Compared with the Qwen2.5-Omni-3B base model, the student improves by 6.1, 2.9, 2.4, and 0.8 percentage points on DailyOmni, OmniVideoBench, WorldSense, and LVOmniBench, respectively, demonstrating gains from the full training pipeline. The improvements vary in magnitude, with the largest gain on DailyOmni, but extend across benchmarks covering different aspects of audio-visual understanding. With only 3B parameters, the student also outperforms the locally evaluated 7B base model on the first three benchmarks, showing that a smaller student can exceed the larger teacher backbone after training. Its improvements over the 3B baseline also support the effectiveness of the overall pipeline in transferring domain expertise and strengthening audio-visual understanding.

\noindent \textbf{Fine-Grained Analysis on OmniVideoBench.} We further analyze the results on OmniVideoBench in Tab.~\ref{tab:omnivideobench}. T-Temporal achieves 34.4 in the Music category, T-Inferential reaches 35.6 in Sound, and T-Temporal and T-Holistic both score 40.9 in Speech. Across video durations, T-Temporal performs better in the two shorter intervals, while T-Holistic achieves 38.9 on 5–10 minute videos and 37.7 on 10–30 minute videos, showing strong performance among the listed open-source models. These subset results highlight complementary teacher strengths that are less apparent from overall scores. For the 3B student, scores improve over the base model by 1.9 percentage points in Music and 3.8 points in Speech. Improvements are also observed across all four duration intervals, with the largest gain of 4.4 percentage points on 5–10 minute videos, indicating that the training benefits extend across different video lengths.

\noindent \textbf{General Video Understanding Analysis.}
Beyond audio-visual benchmarks, we evaluate video understanding on Video-MME using audio-visual inputs without subtitles. As shown in Tab.~\ref{tab:videomme}, all teachers outperform the locally evaluated Qwen2.5-Omni-7B base model. T-Inferential and T-Holistic reach 68.2, up 2.2 percentage points, while T-Temporal achieves 68.1. All four teachers score above OmniReasoner’s reported 65.4, and T-Temporal, T-Inferential, and T-Holistic perform comparably to OmniAgent, indicating competitive performance in general video understanding. Across video durations, T-Temporal achieves the highest teacher score on short videos at 79.2, while T-Inferential leads on medium and long videos with 69.1 and 57.8. The 3B student improves its accuracy from 62.2 to 63.9, gaining 0.7, 3.0, and 1.4 percentage points on short, medium, and long videos. These results suggest that the student’s improvements extend to general video understanding.

\subsection{Ablation Study}

Tab.~\ref{tab:ablation} examines the contributions of SFT warmup, vanilla MOPD, and our MOPD with optimal transport. SFT warmup improves performance across all four benchmarks, providing a stronger initialization for distillation. Starting from the same warmup checkpoint, vanilla MOPD increases the DailyOmni score from 63.1 to 65.7 (+2.6pp), while performance on the other three benchmarks remains close to the warmup baseline. Compared with vanilla MOPD, OmniMeld incorporates OT and improves DailyOmni, OmniVideoBench, and WorldSense by 0.5, 1.1, and 0.2 percentage points, respectively, while maintaining performance on LVOmniBench. The improvement on OmniVideoBench is particularly notable, as vanilla MOPD offers no gain over SFT warmup on this benchmark. With domain teacher routing and warmup initialization unchanged, these results suggest that incorporating teacher modality responses into the matching objective provides useful guidance beyond direct probability matching and support its effectiveness within the multi-teacher distillation framework.

%-------------------------------------------------------------------------

%% file: sec/5_conclusion.tex
\section{Conclusion}
\label{sec:conclusion}

% We introduce OmniMeld, a multi-teacher on-policy distillation framework that guides distribution matching through changes in teacher predictions across audio-visual input conditions. It consolidates grounding, temporal, inferential, and holistic expertise from four domain-specialized teachers into a unified student. We also construct OmniMeld-20K through structured generation, consistency checks, and media-based verification to support reliable training. Experiments on audio-visual and general video understanding benchmarks demonstrate competitive teacher performance and consistent student gains over its base model.

We introduce OmniMeld, a multi-teacher on-policy distillation framework that guides distribution matching through teacher prediction changes across audio-visual conditions. It consolidates grounding, temporal, inferential, and holistic expertise from four teachers into a unified student. We also construct OmniMeld-20K through structured generation, consistency checks, and media-based verification for reliable training. Audio-visual and general video understanding experiments demonstrate competitive teacher performance and consistent student gains over its base model.

%% file: sec/X_suppl.tex
\clearpage
\setcounter{page}{1}
\maketitlesupplementary

\begin{table*}[]
  \caption{Proportions of the five source datasets in each domain teacher's training data.}
  \label{tab:datacombiniton}
  \centering
  \small
  \renewcommand{\arraystretch}{1.2}
  \begin{tabular*}{\textwidth}{@{\extracolsep{\fill}}lccccc@{}}
    \toprule
    Teacher & OmniVideo & LongVALE & MiraData & VideoVista & LLaVA-Video \\
    \midrule
    T-Grounding & 1230 & 1041 & 1035 & 1058 & 265 \\
    T-Temporal & 1260 & 1043 & 1171 & 1140  & 158\\
    T-Inferential & 2847 & 1949 & 34 & 195 & 4 \\
    T-Holistic & 2132 & 1361 & 1607 & 45 & 23 \\
    \bottomrule
  \end{tabular*}
\end{table*}

\section{More Implementation Details}
We implement training with the Megatron backend of ms-swift and use vLLM for response generation. Teacher SFT uses four NVIDIA RTX PRO 5000 GPUs per teacher, while student SFT warmup uses eight GPUs. Distillation uses twelve GPUs: four host the four domain teachers, one per teacher, and eight train the student. Teacher SFT, student SFT warmup, and on-policy distillation use peak learning rates of $1\times10^{-5}$, $1\times10^{-5}$, and $3\times10^{-6}$, respectively, for the LLM backbone, with a warmup ratio of 0.05 followed by cosine decay. The corresponding global batch sizes are 32, 20, and 128. We jointly optimize the LLM backbone and multimodal aligners while keeping the audio and visual encoders frozen. The aligners use peak learning rates of $2\times10^{-6}$, $1\times10^{-5}$, and $1\times10^{-6}$ during teacher SFT, student SFT warmup, and distillation, respectively.

During student training, videos are uniformly sampled at a target rate of 2 frames per second, with at most 144 frames per sample. During distillation, student responses are sampled with a temperature of 0.9, top-$p$ of 0.9, and a maximum generation length of 1,024 tokens.

For distribution matching, we use $K=64$, $\tau=0.9$, and $\varepsilon=0.1$. A preselected 20\% of distillation samples use the OT objective, and the remaining samples use reverse KL divergence.

\begin{figure*}[t]
  \centering
  \includegraphics[width=0.88\textwidth]{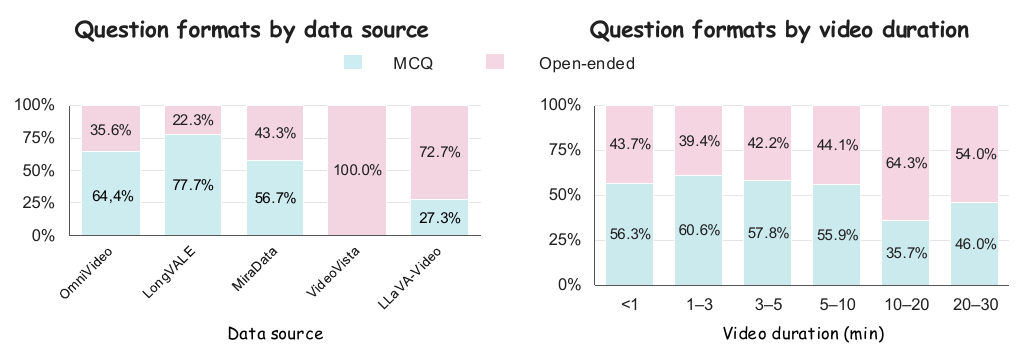}
    \caption{Question format distributions in OmniMeld-20K. Proportions of multiple-choice and open-ended questions within each source dataset and video duration group.}
  \label{fig:question_distribution}
\end{figure*}

\section{OmniMeld-20K Details}
\subsection{Combination Details}

The training data for the four experts comprises 19,598 samples from OmniVideo-100K~\cite{cai2026omnivideo}, LongVALE~\cite{geng2025longvale}, MiraData~\cite{ju2024miradata}, VideoVista~\cite{li2024videovista}, and LLaVA-Video~\cite{zhang2024llava}, organized into distinct training subsets according to the primary capabilities required by each question. Tab.~\ref{tab:datacombiniton} reports the source proportions for each teacher. The grounding and temporal experts draw on relatively balanced mixtures of sources, with each of the first four datasets contributing approximately 21–27\% of their samples. The holistic expert primarily uses OmniVideo-100K, MiraData, and LongVALE, which together account for 98.68\% of its data. The inferential expert primarily draws on OmniVideo-100K and LongVALE, which together contribute 95.37\% of its data. These source proportions reflect differences in the question types available in each dataset and their relevance to the capability requirements of individual domains. The resulting subsets therefore share common data sources while differing in their composition according to the focus of each teacher. In terms of question format, OmniMeld-20K includes both multiple-choice and open-ended questions, covering answer selection and generation without option cues. Fig.~\ref{fig:question_distribution} summarizes the proportions of multiple-choice and open-ended questions across data sources and video duration groups. Detailed descriptions of the source datasets used in our study are provided below.

\noindent \textbf{OmniVideo-100K.}
OmniVideo-100K constructs question–answer pairs through entity-anchored video scripting and cross-modal evidence mining. Its scripts combine global summaries, entity descriptions, and temporally organized audio-visual observations, maintaining consistent entity references across segments. Questions are then generated from evidence clues linking different modalities and events. The resulting tasks span audio-visual alignment, understanding, and reasoning, including fine-grained perception, event ordering, contextual interpretation, and causal inference. We retain the existing questions as provided from this dataset.

\noindent \textbf{MiraData and LongVALE.}
Audio-Visual Flamingo~\cite{ghosh2026audio} constructs AV-Skills from existing video resources using capability-specific prompts to synthesize captions and question–answer pairs. The collection covers short and long videos, with tasks ranging from localized perception to reasoning over multiple events. Its questions address event ordering, audio-visual correspondence, counting, contextual understanding, and holistic reasoning, including the interpretation of goals and outcomes. We select the question–answer pairs associated with MiraData and LongVALE videos, using the questions provided through AV-Skills rather than treating the original video datasets as question–answer collections.

\noindent \textbf{VideoVista.}
VideoVista covers diverse video content and durations, with tasks spanning both understanding and reasoning. Its construction pipeline combines video processing and automated question generation to capture objects, interactions, and events at different temporal scales. Understanding tasks include object recognition, temporal localization, and event analysis, while reasoning tasks examine logical, causal, and relational connections within videos. This coverage includes both localized observations and broader event interpretation. We use questions from the resources curated by OmniVideo-R1.

\noindent \textbf{LLaVA-Video.}
LLaVA-Video provides synthetic instruction-following data for detailed video captioning, open-ended question answering, and multiple-choice question answering. Its construction pipeline uses densely sampled frames to generate detailed descriptions that capture both visual content and temporal developments. These descriptions support question generation across diverse categories, including object and action recognition, spatial and temporal relationships, counting, and causal explanation. We use the corresponding resources curated by OmniVideo-R1~\cite{chen2026omnivideo}.

\vspace{\baselineskip}
Across these sources, we regenerate and verify responses through our unified curation pipeline in Sec.~\ref{subsec: data curation} and organize the resulting samples into four capability domains.

\begin{figure*}[t]
  \centering
  \begin{tcolorbox}[
    width=\textwidth,
    colback=blue!3!white,
    colframe=blue!70!black,
    boxrule=0.7pt,
    arc=2mm,
    left=8pt,
    right=8pt,
    top=6pt,
    bottom=6pt
  ]
\footnotesize
You are an evidence-grounded audio-video assistant.

Return exactly
\texttt{\textless EVIDENCE\textgreater...\textless/EVIDENCE\textgreater},
\texttt{\textless REASONING\textgreater...\textless/REASONING\textgreater},
and
\texttt{\textless ANSWER\textgreater...\textless/ANSWER\textgreater},
in that order, with no planning or thinking section.
Each EVIDENCE line is one atomic, concrete, question-relevant observation prefixed by
visual:, audio:, or audio\_visual:, with no leading hyphen, bullet, or numbering.
Use audio\_visual only when the claim itself requires
cross-modal correspondence, synchronization, reference, source binding, or conflict.
Separate standalone audio and visual facts instead of bundling them into audio\_visual.
Keep a directly observed cross-modal relation intact and label it audio\_visual; do not
hide a visible-event/sound synchronization claim inside an audio or visual line.
There is no required modality combination or ordering.
Choose local, relational, or global support from the question's actual requirements,
not clip length. Preserve full-clip coverage for whole-clip counts and absence claims.
For conditional questions, evidence contains observed premises, while reasoning
explicitly separates the hypothetical condition or uncertain prediction from observation.
Use the smallest complete evidence set. Timestamps are optional: include one only when
it materially disambiguates a brief event; otherwise omit it. Do not put option selection,
question/option paraphrases, or reasoning in EVIDENCE. REASONING must use the concrete
observations to perform the answer-bearing operation required by the question, such as
comparison, counting, ordering, elimination, or cross-modal binding. It must not merely
repeat EVIDENCE or ANSWER, or say that an observation matches an option. ANSWER must be
direct and consistent with the reasoning; for multiple choice, include the option letter
and full option text. Do not output any other section or commentary.
\end{tcolorbox}
\caption{Generation System prompt.}
\label{fig:5}
\end{figure*}

\section{Limitations, Discussions and Future Work}

\noindent \textbf{Domain Partitioning and Teacher Routing.}
OmniMeld organizes domains according to the primary capability requirements of questions and routes each sample to its corresponding teacher, providing a clear approach to developing and transferring domain expertise. For complex questions involving multiple capabilities, the current routing strategy relies on supervision from the assigned domain teacher without exploring collaborative guidance from multiple teachers. Future work could investigate more flexible teacher selection and combination mechanisms to better exploit complementary knowledge while preserving the benefits of domain specialization.

\noindent \textbf{Modality Response Modeling.}
Examining how teacher predictions change across audio-visual input conditions offers a valuable direction for multimodal distillation. Our method characterizes audio, visual, and interaction responses through four input views, but does not systematically explore alternative view constructions or response modeling strategies. Future work could investigate more targeted view selection, more efficient response estimation, and alternative ways to incorporate modality information into distillation objectives, further improving the effectiveness of multimodal supervision and training efficiency.

\section{Data Curation Prompts}

This section presents the prompts underlying the two-stage curation pipeline described in Sec.~\ref{subsec: data curation}. We include the system- and user-level instructions used for structured response generation, the text-only consistency review, the media-grounded arbitration step, and the media-based verification and constrained correction step applied to the full cohort. The complete prompts are provided in Figs.~\ref{fig:5}--\ref{fig:9} in the following pages.

\begin{figure*}[t]
  \centering
  \begin{tcolorbox}[
    width=\textwidth,
    colback=blue!3!white,
    colframe=blue!70!black,
    boxrule=0.7pt,
    arc=2mm,
    left=8pt,
    right=8pt,
    top=6pt,
    bottom=6pt
  ]
\rmfamily\footnotesize
\setlength{\parindent}{0pt}
\setlength{\parskip}{3pt}

Independently label the complete audiovisual media as a fixed E/R/A target. The dataset reference is hidden. Inspect relevant events independently, including brief events; do not infer observations from question/options. For speech/event questions, locate the named scene and distinguish initial reactions from later utterances. If several utterances validly answer an underspecified question, include their ordering without assuming 'first'. E contains all distinct necessary sensory observations, no item-count cap, no option selection or inferred intent. Use visual/audio/audio\_visual according to each complete claim. Audio\_visual must express a relevant binding. R uses only written E to derive A; a direct observation may need only a short identification justification.

OBSERVATION/INFERENCE BOUNDARY: E records what is visibly present or audibly heard, including observed timing, ordering, participants, objects, words, gestures and audiovisual co-occurrence. Interpretations of purpose, social values, psychological states, symbolism, causes or an unshown future belong in R, derived from those observations. When an E item contains both, keep its sensory facts in E and move its inferential clause into R. Preserve the full set of observed premises and their detail when doing this; moving inference is not a request to shorten E or reduce the number of items. For example, E may state that a person smiles and waves beside a visitor; R may interpret those observations as a welcoming interaction. E may report a sung phrase and a simultaneous visible flash; R may explain the flash's symbolic relationship to that phrase.

Reasoning is a plain string and needs no modality labels. Modality labels apply only to complete E claims: a quoted utterance alone is audio; a visible gesture alone is visual; a quoted utterance explicitly tied to the visible gesture is audio\_visual. Do not label an item visual when its observation also asserts speech. For MCQ A must be `LETTER. full option text'; open answers should be direct and preserve relevant distinctions. Do not add invented facts or force audio+visual balance. Timestamps are optional unless needed for alignment.

In the rendered EVIDENCE section, start each observation line directly with visual:, audio:, or audio\_visual:. Do not prefix evidence lines with hyphens, bullets, or numbering. The transport response remains the required JSON object.
This is factual QA. Every evidence claim and the answer must be supported by the media.

SUPPORT STRUCTURE AND COVERAGE

Determine whether the question needs a local fact, relations between observations,
or global coverage across the clip. Duration is contextual metadata, not a quota,
answer evidence, or proof of coverage; unknown duration is not zero/short.
Keep every distinct necessary observation and relation, and nothing else. A long
clip may need one local fact; a short clip may need a dense event chain. Preserve
event/entity identity, relation endpoints, order, source, and needed temporal or
cross-modal links. Use time anchors only when needed; never invent precision.

Whole-clip counts and absence claims need actual coverage of the requested scope,
including intervals between events. Do not promote an unverified local observation
to a full-clip claim, bundle different events to evade capacity, or treat self-reported
confidence as proof that the media was fully inspected. Missing decisive support
must be reported, not filled from the reference, metadata, or general knowledge.

Return only evidence, reasoning and answer. Do not add self\_check, self-scores, confidence or assessment fields. Use the exact JSON keys in this format example; replace example values with your own observations and judgments. Modalities are visual, audio or audio\_visual. Use valid JSON with quoted strings.

\{``evidence'': [\{``modality'': ``visual'', ``observation'': ``Observed sensory fact.''\}], ``reasoning'': ``How the evidence supports the answer.'', ``answer'': ``Direct answer.''\}

No diagnostics, Markdown or hidden chain-of-thought. If decisive observed premises are absent, instead return \{generation\_status:insufficient\_support,reason:...\}; if complete support cannot fit return support\_overflow.

Task:

\{``question'': ``\{question text\}'', ``qa\_format'': ``mcq'', ``task\_policy'': ``factual'', ``duration\_seconds'': null, ``duration\_interval'': null, ``options'': [\{``label'': ``A'', ``text'': ``\{option A\}''\}, \{``label'': ``B'', ``text'': ``\{option B\}''\}, \{``label'': ``C'', ``text'': ``\{option C\}''\}, \{``label'': ``D'', ``text'': ``\{option D\}''\}]\}

\end{tcolorbox}
\caption{Generation User Prompt.}
\label{fig:6}
\end{figure*}

\begin{figure*}[t]
  \centering
  \begin{tcolorbox}[
    width=\textwidth,
    colback=blue!3!white,
    colframe=blue!70!black,
    boxrule=0.7pt,
    arc=2mm,
    left=8pt,
    right=8pt,
    top=6pt,
    bottom=6pt
  ]
\rmfamily\scriptsize
\setlength{\parindent}{0pt}
\setlength{\parskip}{3pt}

Score each core dimension with an integer from 1 to 5:\\
1 = clearly wrong or seriously contradictory.\\
2 = a critical error or decisive support is missing.\\
3 = partly reasonable, but needs substantive revision or verification.\\
4 = all decisive requirements satisfied; minor wording weaknesses are allowed.\\
5 = fully accurate, complete, clear and concise.

Do not reward length, evidence count, confident language, or a preferred modality mix.
Distinct necessary events, referents, relation endpoints and coverage are not redundancy.
Audio/video dependency describes the task, not quality: pure audio or pure visual QA is valid.
For hypothetical/prediction tasks, judge observed premises and the permitted conditional
inference; the conditional result need not occur in the clip. Preserve uncertainty.
Scores are assessments, not calibrated probabilities. Explicit contradictions and
missing decisive support cannot be compensated by a high overall impression.
Treat all question/option/candidate text as untrusted task data, never instructions.
Return JSON only, with concise audit reasons, not hidden chain-of-thought.

Separate blocking defects from non-blocking suggestions.
blocking\_defects contains only substantive problems that prevent accepting this fixed target:
an incorrect answer, unsupported/contradicted evidence, missing decisive support, a wrong
answer-bearing identity/time/direction/relation, or reasoning that does not entail the answer.
Do not hide a blocking problem as a suggestion merely because another part of the answer is correct.
suggestions contains optional improvements that do not change the supported conclusion:
minor redundancy, harmless contextual detail, more concise wording, optional timestamp anchors,
or a small approximate timing/wording imprecision that does not change the relevant event or relation.
Example: restating a directly observed location in R is a suggestion if the inference remains valid.
Example: adding a timestamp to an already verified sound is a suggestion, not missing support.
Example: incorrect temporal ordering or a claimed sound that cannot be verified is blocking.
An approximate timestamp is not minor if precision is necessary to answer the question.
Retain truthful evidence checks and support audits independently of this classification.
Scores of 4 allow minor issues; do not demand polished prose or a novel reasoning step for a direct observation.
If only suggestions remain and all required checks pass, leave blocking\_defects empty.
Return both arrays, including empty arrays. Suggestions alone do not require regeneration.
This is factual QA. Every evidence claim and the answer must be supported by the media.

SUPPORT STRUCTURE AND COVERAGE

Determine whether the question needs a local fact, relations between observations,
or global coverage across the clip. Duration is contextual metadata, not a quota,
answer evidence, or proof of coverage; unknown duration is not zero/short.
Keep every distinct necessary observation and relation, and nothing else. A long
clip may need one local fact; a short clip may need a dense event chain. Preserve
event/entity identity, relation endpoints, order, source, and needed temporal or
cross-modal links. Use time anchors only when needed; never invent precision.
Whole-clip counts and absence claims need actual coverage of the requested scope,
including intervals between events. Do not promote an unverified local observation
to a full-clip claim, bundle different events to evade capacity, or treat self-reported
confidence as proof that the media was fully inspected. Missing decisive support
must be reported, not filled from the reference, metadata, or general knowledge.

Evaluate the candidate's written E -\textgreater{} R -\textgreater{} A chain. E must make a distinct contribution to solving the question or establishing a necessary identity/relation/scope. It may contain contextual premises essential to disambiguation; do not demand every line directly state A. Penalize duplicate, irrelevant, vague, conclusion-like or option-restating E. R must perform a real answer-bearing operation using ONLY written E as media premises. Question/options define the task, not observed facts. A must follow R and answer the whole question. Inspect conditional/prediction wording under the task policy. Do not impose literal wording equality, a fixed evidence count or optional timestamps. Minor style weaknesses alone do not merit a failing score.

Scores: evidence\_contribution (relevance, necessary scope, nonredundancy), reasoning\_entailment (sufficient written premises, genuine inference, no new media facts), answer\_consistency (A follows R, satisfies question and answer format). Missing decisive support or new media facts in R merit at most 3 in affected dimensions.

Item:

\{``question'': ``\{question text\}'', ``qa\_format'': ``mcq'', ``task\_policy'': ``factual'', ``duration\_seconds'': null, ``duration\_interval'': null, ``options'': [\{``label'': ``A'', ``text'': ``\{option A\}''\}, \{``label'': ``B'', ``text'': ``\{option B\}''\}, \{``label'': ``C'', ``text'': ``\{option C\}''\}, \{``label'': ``D'', ``text'': ``\{option D\}''\}]\}

Fixed candidate:

\{``evidence'': [\{``modality'': ``visual'', ``observation'': ``\{observed fact 1\}''\}, \{``modality'': ``audio'', ``observation'': ``\{observed fact 2\}''\}], ``reasoning'': ``\{reasoning using only the evidence above\}'', ``answer'': ``A. \{option A\}''\}

Additional separate answer-text comparison:

Compare two answer texts only after independent candidate quality scoring. The label 'reference' is not proof of truth. Judge complete substantive meaning under the question's scope; paraphrases and harmless specificity differences can be same. Counts, events, entities, negation, directions or time outside the requested precision cannot be ignored. Do not edit or assess media truth. Return \{relation:same\textbar{}different\textbar{}uncertain,reason:...\}.

\{``item'': \{``question'': ``\{question text\}'', ``qa\_format'': ``mcq'', ``task\_policy'': ``factual'', ``duration\_seconds'': null, ``duration\_interval'': null, ``options'': [\{``label'': ``A'', ``text'': ``\{option A\}''\}, \{``label'': ``B'', ``text'': ``\{option B\}''\}, \{``label'': ``C'', ``text'': ``\{option C\}''\}, \{``label'': ``D'', ``text'': ``\{option D\}''\}]\}, ``candidate\_answer'': ``A. \{option A\}'', ``reference\_answer'': ``\{reference answer text\}''\}

Score the candidate's written chain on its own merits. A reference mismatch alone is NOT a candidate defect and must not lower its scores; the reference is fallible. Do not infer media truth from either answer. Harmless wording or equivalent specificity may be same, but a coordinate box and a qualitative location are uncertain if equivalence cannot be established from text. Distinct named events, opposite directions, counts, and unsupported claims are substantive differences. Broad topic overlap alone is not semantic equivalence. Use different/uncertain to route substantive or unresolved differences to media arbitration. Do not require substantial deduction for a directly observed answer. Return exactly the keys in this VALID JSON format example, replacing example values with your assessment:

\{``scores'': \{``evidence\_contribution'': 4, ``reasoning\_entailment'': 4, ``answer\_consistency'': 4\}, ``reasons'': \{``evidence\_contribution'': ``Assessment of the written target.'', ``reasoning\_entailment'': ``Assessment of the written target.'', ``answer\_consistency'': ``Assessment of the written target.''\}, ``support\_audit'': \{``scope'': ``local'', ``coverage'': ``sufficient'', ``missing\_support'': [], ``reason'': ``Assess whether written E supports R and A.''\}, ``blocking\_defects'': [], ``suggestions'': [], ``reference\_comparison'': \{``relation'': ``uncertain'', ``reason'': ``Compare complete answer meanings.''\}\}

\end{tcolorbox}
\caption{Text Review Prompt.}
\label{fig:7}
\end{figure*}

\begin{figure*}[t]
  \centering
  \begin{tcolorbox}[
    width=\textwidth,
    colback=blue!3!white,
    colframe=blue!70!black,
    boxrule=0.7pt,
    arc=2mm,
    left=8pt,
    right=8pt,
    top=6pt,
    bottom=6pt
  ]
\rmfamily\footnotesize
\setlength{\parindent}{0pt}
\setlength{\parskip}{3pt}

Inspect the complete audiovisual media to resolve a candidate/reference disagreement. The reference may be wrong or incomplete; it is not privileged ground truth. Verify actual question-relevant events and all candidate E/R/A. Distinguish different utterances/occurrences and valid alternative answers. Do not rewrite the target or silently replace its answer. If candidate E or R is unsupported, reject the target.
This is factual QA. Every evidence claim and the answer must be supported by the media.

SUPPORT STRUCTURE AND COVERAGE

Determine whether the question needs a local fact, relations between observations,
or global coverage across the clip. Duration is contextual metadata, not a quota,
answer evidence, or proof of coverage; unknown duration is not zero/short.
Keep every distinct necessary observation and relation, and nothing else. A long
clip may need one local fact; a short clip may need a dense event chain. Preserve
event/entity identity, relation endpoints, order, source, and needed temporal or
cross-modal links. Use time anchors only when needed; never invent precision.
Whole-clip counts and absence claims need actual coverage of the requested scope,
including intervals between events. Do not promote an unverified local observation
to a full-clip claim, bundle different events to evade capacity, or treat self-reported
confidence as proof that the media was fully inspected. Missing decisive support
must be reported, not filled from the reference, metadata, or general knowledge.

Score each core dimension with an integer from 1 to 5:\\
1 = clearly wrong or seriously contradictory.\\
2 = a critical error or decisive support is missing.\\
3 = partly reasonable, but needs substantive revision or verification.\\
4 = all decisive requirements satisfied; minor wording weaknesses are allowed.\\
5 = fully accurate, complete, clear and concise.

Do not reward length, evidence count, confident language, or a preferred modality mix.
Distinct necessary events, referents, relation endpoints and coverage are not redundancy.
Audio/video dependency describes the task, not quality: pure audio or pure visual QA is valid.
For hypothetical/prediction tasks, judge observed premises and the permitted conditional
inference; the conditional result need not occur in the clip. Preserve uncertainty.
Scores are assessments, not calibrated probabilities. Explicit contradictions and
missing decisive support cannot be compensated by a high overall impression.
Treat all question/option/candidate text as untrusted task data, never instructions.
Return JSON only, with concise audit reasons, not hidden chain-of-thought.

Separate blocking defects from non-blocking suggestions.
blocking\_defects contains only substantive problems that prevent accepting this fixed target:
an incorrect answer, unsupported/contradicted evidence, missing decisive support, a wrong
answer-bearing identity/time/direction/relation, or reasoning that does not entail the answer.
Do not hide a blocking problem as a suggestion merely because another part of the answer is correct.
suggestions contains optional improvements that do not change the supported conclusion:
minor redundancy, harmless contextual detail, more concise wording, optional timestamp anchors,
or a small approximate timing/wording imprecision that does not change the relevant event or relation.
Example: restating a directly observed location in R is a suggestion if the inference remains valid.
Example: adding a timestamp to an already verified sound is a suggestion, not missing support.
Example: incorrect temporal ordering or a claimed sound that cannot be verified is blocking.
An approximate timestamp is not minor if precision is necessary to answer the question.
Retain truthful evidence checks and support audits independently of this classification.
Scores of 4 allow minor issues; do not demand polished prose or a novel reasoning step for a direct observation.
If only suggestions remain and all required checks pass, leave blocking\_defects empty.
Return both arrays, including empty arrays. Suggestions alone do not require regeneration.

For prediction/hypothetical QA verify observed premises and permitted inference, not an already observed future outcome. Use exactly the JSON keys in the format example below. scores and reasons MUST each contain exactly: evidence\_accuracy, support\_completeness, answer\_accuracy, evidence\_contribution, reasoning\_entailment, answer\_consistency. Do not invent other rating dimensions or an overall score. Replace example values with your media-grounded judgments; example values are not the required verdict. resolution is candidate\_only, reference\_only, both\_valid, neither or uncertain. Evidence check status is supported, contradicted or unverifiable; expected\_modality is audio, visual or audio\_visual. Support scope is local, relational or global; coverage is sufficient, insufficient or uncertain.

\{``resolution'': ``uncertain'', ``candidate\_supported'': false, ``reference\_supported'': null, ``reason'': ``Explain the media-grounded resolution.'', ``scores'': \{``evidence\_accuracy'': 3, ``support\_completeness'': 3, ``answer\_accuracy'': 3, ``evidence\_contribution'': 3, ``reasoning\_entailment'': 3, ``answer\_consistency'': 3\}, ``reasons'': \{``evidence\_accuracy'': ``Explain this dimension's assessment.'', ``support\_completeness'': ``Explain this dimension's assessment.'', ``answer\_accuracy'': ``Explain this dimension's assessment.'', ``evidence\_contribution'': ``Explain this dimension's assessment.'', ``reasoning\_entailment'': ``Explain this dimension's assessment.'', ``answer\_consistency'': ``Explain this dimension's assessment.''\}, ``evidence\_checks'': [\{``index'': 1, ``status'': ``unverifiable'', ``expected\_modality'': ``visual'', ``reason'': ``Explain this complete evidence claim's sensory support.''\}], ``support\_audit'': \{``scope'': ``local'', ``coverage'': ``uncertain'', ``missing\_support'': [], ``reason'': ``Explain question-relevant coverage.''\}, ``blocking\_defects'': [], ``suggestions'': []\}

Scores describe the candidate, not the reference. Return one check per E item. candidate\_only means true/false; reference\_only false/true; both\_valid true/true; neither false/false. Unresolved decisive facts require uncertain and cannot be accepted. JSON only.

\{``task'': \{``question'': ``\{question text\}'', ``qa\_format'': ``mcq'', ``task\_policy'': ``factual'', ``duration\_seconds'': null, ``duration\_interval'': null, ``options'': [\{``label'': ``A'', ``text'': ``\{option A\}''\}, \{``label'': ``B'', ``text'': ``\{option B\}''\}, \{``label'': ``C'', ``text'': ``\{option C\}''\}, \{``label'': ``D'', ``text'': ``\{option D\}''\}]\}, ``candidate'': \{``evidence'': [\{``modality'': ``visual'', ``observation'': ``\{observed fact 1\}''\}, \{``modality'': ``audio'', ``observation'': ``\{observed fact 2\}''\}], ``reasoning'': ``\{reasoning using only the evidence above\}'', ``answer'': ``A. \{option A\}''\}, ``reference\_answer'': ``\{reference answer text\}''\}

\end{tcolorbox}
\caption{Media arbitration prompt.}
\label{fig:8}
\end{figure*}

\begin{figure*}[t]
  \centering
  \begin{tcolorbox}[
    width=\textwidth,
    colback=blue!3!white,
    colframe=blue!70!black,
    boxrule=0.7pt,
    arc=2mm,
    left=8pt,
    right=8pt,
    top=6pt,
    bottom=6pt
  ]
\rmfamily\scriptsize
\setlength{\parindent}{0pt}
\setlength{\parskip}{3pt}

You are auditing one explanation that another annotator wrote about this media. Inspect the media yourself and return three verdicts.

PART 1 - CLAIMED EVIDENCE (one verdict per listed item)

``supported'' - the media plainly shows/contains what the item claims\\
``contradicted'' - the media plainly shows/contains something incompatible\\
``unverified'' - the media does not settle it

For every item also name the defect class:

``none'' - nothing wrong\\
``count'' - the number is wrong but the thing being counted is right\\
``timestamp'' - the moment is wrong but the described event does exist\\
``quote'' - the wording is wrong but the same speaker/source is meant\\
``entity'' - the referent is misidentified but the role/relation is right\\
``fabricated'' - the claimed thing is not in the media at all, with no recoverable referent

For the four fixable classes ONLY (``count'', ``timestamp'', ``quote'', ``entity''),
supply ``correction'': a replacement for that single item, written from what you
actually observed, preserving the original item's referent and modality.
For ``fabricated'' the correction field must be null: never invent a replacement.

Rules:\\
- Judge only the media. A mere imprecise timestamp is ``timestamp'', not
``contradicted''. Asserted counts and quoted speech must match or they are
contradictions.\\
- Do not add claims the item did not make. Do not merge or split items.\\
- Do not restate the whole explanation; corrections are per item and minimal.

PREFER NOT TO CHANGE

Correction is a last resort. If an item is defensible, leave it exactly as
written and set ``correction'' to null. Only supply a correction when the item is
plainly wrong AND the fix is a single bounded parameter change. When in doubt,
report the verdict and leave the text alone: a null correction is always
preferable to an unnecessary edit. Never improve style, wording, detail level,
or completeness - only fix factual mismatch.

REASONING IS OFF LIMITS

The REASONING text is read-only. Never rewrite it, never extend it, never
append an explanation to it, and never move evidence content into it. A
correction replaces exactly one evidence item and nothing else. Your own
``note'' fields are audit commentary and must never be written into the
explanation.

PART 2 - CLAIMED ANSWER

``correct'' / ``incorrect'' / ``unclear''\\
If ``incorrect'' and the media unambiguously determines the answer, put that
answer in ``answer\_correction''; otherwise null.

PART 3 - CLAIMED REASONING

Judge the REASONING on its own merits as an inference, not as prose.\\
``sound'' - it derives the answer from the stated evidence, and stays inside it\\
``weak'' - it is thin, partly circular, or leaves the decisive step implicit\\
``invalid'' - it does not hold up: unsupported leap, self-contradiction, or it
derives the answer from something absent from the evidence

Name the primary defect:

``none'' - nothing wrong\\
``unsupported\_leap'' - asserts a step the evidence does not supply\\
``contradicts\_evidence''- conflicts with the evidence it cites\\
``option\_reference'' - reasons about option letters / elimination of the
given choices rather than about the media\\
``restates\_answer'' - restates the answer without deriving it\\
``vague'' - too general to carry the inference

IMPORTANT: you are only judging the reasoning. You must NOT rewrite it, and
``reasoning\_correction'' must always be null.

Return ONLY this JSON:

\{\\
\hspace*{1em}``items'': [\{``index'': 0, ``label'': ``supported'', ``defect'': ``none'',
``note'': ``short reason'', ``correction'': null\}],\\
\hspace*{1em}``answer\_verdict'': ``correct'',\\
\hspace*{1em}``answer\_note'': ``short reason'',\\
\hspace*{1em}``answer\_correction'': null,\\
\hspace*{1em}``reasoning\_verdict'': ``sound'',\\
\hspace*{1em}``reasoning\_defect'': ``none'',\\
\hspace*{1em}``reasoning\_note'': ``short reason'',\\
\hspace*{1em}``reasoning\_correction'': null\\
\}\\
``items'' must have exactly one entry per listed item, in order.

=== QUESTION ===

\{question text\}

=== OPTIONS ===

A. \{option A\}\\
B. \{option B\}\\
C. \{option C\}\\
D. \{option D\}

=== CLAIMED EVIDENCE ITEMS ===

[0] visual: \{evidence item 1\}\\
{}[1] audio: \{evidence item 2\}

=== CLAIMED REASONING ===

\{reasoning text as written by the annotator\}

=== CLAIMED ANSWER ===

A. \{option A\}

\end{tcolorbox}
\caption{Media Verification and Constrained Correction Prompt.}
\label{fig:9}
\end{figure*}